\documentclass[11pt,a4paper]{article}
\usepackage[hyperref]{eacl2021}
\usepackage{times}
\usepackage{latexsym}

\usepackage{hyperref}
\usepackage{graphicx}
\usepackage{tabularx}
\usepackage{subcaption}
\usepackage{framed}
\usepackage{xcolor}
\usepackage{verbatim}

\usepackage{microtype}

\aclfinalcopy % Uncomment this line for the final submission
\usepackage{todonotes}

\title{``Sharks are not the threat humans are'': Argument Component Segmentation in School Student Essays}

\author{Tariq Alhindi \thanks{\hspace{.12cm}Work performed during internship at ETS}\\
  Department of Computer Science \\
  Columbia University \\
  \texttt{tariq@cs.columbia.edu} \\\And
  Debanjan Ghosh  \\
  Educational Testing Service \\
  \texttt{dghosh@ets.org} \\}

\date{}

\begin{document}
\maketitle

\begin{abstract}
Argument mining is often addressed by a pipeline method where segmentation of text into argumentative units is conducted first and proceeded by an argument component identification task. In this research, we apply a token-level classification to identify claim and premise tokens from a new corpus of argumentative essays written by middle school students. To this end, we compare a variety of state-of-the-art models such as discrete features and deep learning  architectures  (e.g., BiLSTM networks and BERT-based architectures) to identify the argument components. We demonstrate that a BERT-based multi-task learning architecture (i.e., token and sentence level classification) adaptively pretrained on a relevant unlabeled dataset obtains the best results.
\end{abstract}

%===============================================%
%===============================================%
%===============================================%

% comments from reviews that we can address
\begin{comment}
% Reviewer 1:

% Reviewer 2:
W1: The idea to jointly segment and classify components (on the token level) is sold as a bit more new than it actually is to me.

W2: The weak performance of the approaches on premises should be discussed more in detail.

W3: Significant tests are missing to clarify which differences are actual improvements. Moreover, a real comparison to previous approaches from other works is left out.

Q1: Table 1: It seems each paragraph has one sentence only here. Is this the case, or is it just for display purposes?

201 [Q3]: The paper argues only what makes schools student essays more challenging than other student essays. I wondered whether there may be also aspects in which they are less challenging?

Table 3: I would recommend to also explicitly list a trivial majority-class baseline for comparison. This allows seeing directly how much the models actually learn. If I do the math right based on Table 2, it should have an accuracy of 0.600, i.e., more than Discrete*.

% Reviewer 3:

\end{comment}

%===============================================%
%===============================================%
%===============================================%

\section{Introduction}
\label{sec:intro}

% \begin{figure}[t]
% \centering
% % \includegraphics[width=0.9\linewidth, height=10cm]{figures/figure-sweetners.pdf}
% \includegraphics[width=\linewidth, height=11cm]{figures/figure-sweetners.pdf}
% \caption{Excerpt from one annotated essay with Claim and Premise segments and relations between segments. \textbf{B-Claim} and \textbf{B-Premise} are bold in the Figure.}
% \label{fig:tagged_intro}
% \end{figure}

%Argumenttion requires the ability to come up with  claims and their supports based on relevant examples expressing authors' perspective on a topic.

Computational argument mining focuses on subtasks such as identifying the Argumentative Discourse Units (ADUs) \cite{peldszus2013argument}, their nature (i.e., claim or premise), and the relation (i.e., support/attack) between them \cite{ghosh2014analyzing,wacholder2014annotating,stab2014identifying,stab2017parsing,stede2018argumentation,nguyen2018argument,lawrence2020argument}. Argumentation is essential in academic writing as it enhances the logical reasoning, as well as, critical thinking capacities of students \cite{ghosh2020exploratory}. Thus, in recent times, argument mining has been used to assess students' writing skills in essay scoring and provide feedback on the writing \cite{song2014applying,somasundaran2016evaluating, wachsmuth-etal-2016-using,zhang-litman-2020-automated}.
%persing2015modeling

\begin{figure}
\centering
\setlength{\tabcolsep}{15pt}%horizontal margin
%\small
%\begin{tabular}{ |l|p{10cm}| } 
    \scalebox{0.7}{

\begin{tabular}{ |p{9cm}| }
\hline
\\
Should Artificial Sweeteners be Banned in America?\\
\\
\color{blue} Diet \color{violet} soda , sugar - free gum, and low - calorie sweeteners are what most people see as a way to sweeten up a day without the calories. \\
\\
%As many Americans seek slimmer bodies or healthier lifestyles, artificial sweeteners are the go - to replacement for sugar.\\
%\\
Despite the lack of calories, \color{blue} artificial \color{violet} sweeteners have multiple negative health effects. \\
 \\
Over the past century, science has made it possible to replicate food with fabricated alternatives that simplify weight loss.\\ 
\\
Although many thought \color{blue} these \color{violet} new replacements would benefit overall health, \color{olive} there \color{teal} are more negative effects on manufactured food than the food they replaced.\\
\\
Artificial sweeteners have a huge impact on current day society.\\
\color{black}\rule{\linewidth}{0.2mm} 
\\
Legends: \color{blue} B-Claim \color{violet} I-Claim \color{olive} B-Premise 
\color{teal} I-Premise \color{black} O-Arg \\\\
\hline

%\multicolumn{1}{|c|}{Platform} & {c}{Turn Type} & 
%\multicolumn{1}{c|}{Turn pairs} \\
\end{tabular}
}
\caption{Excerpt from an annotated essay with Claim  Premise segments in BIO notation} %(\cite{song2017examining}, p. 6).}
\label{figure:tagged_intro}
\end{figure}

While argument mining literature has addressed students writing in the educational context, so far, it has primarily addressed college level writing \cite{blanchard2013toefl11,persing2015modeling,beigman_klebanov_detecting_2017,eger2017neural} except for a very few ones \cite{Attali_Burstein_2006,lugini2018annotating,correnti2020automated}. Instead, in this paper, we concentrate on identifying arguments from essays written by \emph{middle school students}. To this end, we perused a new corpus of 145 argumentative essays written by middle school students to identify the argument components. These essays are obtained from an Educational app - \emph{Writing Mentor} - that operates as a Google-docs Add-on.\footnote{\href{https://mentormywriting.org}{https://mentormywriting.org}} 

Normally, research that investigates college students writing in the context of argument mining apply 
a pipeline of subtasks to first detect arguments at the token-level text units (i.e., argumentative tokens or not), and subsequently classify the text units to argument components \cite{stab2017parsing}. However, middle school student essays are vastly different from college students' writing (detailed in Section \ref{sec:data}). We argue they are more difficult to analyze through the pipeline approach due to run-on sentences, unsupported claims, and the presence of several claims in a sentence. Thus, instead of segmenting the text into argumentative/non-argumentative units first, we conduct a token-level classification task to identify the type of the argument component  (e.g., B/I tokens from claims and premises) directly by joining the first and the second subtask in a single task. Figure \ref{figure:tagged_intro} presents an excerpt from an annotated essay with their corresponding gold annotations of claims (e.g., ``Diet soda \dots the calories'')  and premises (e.g., ``there are \dots replaced''). The legends represent the tokens by the standard BIO notations.

%not only we conduct the first subtask of separating argumentative and non-argumentative tokens

%Claim detection has been mostly approached in previous work as a sentence classification task \cite{daxenberger2017essence, chakrabarty2019imho}, or done at two steps at the token level by first finding claims merged with other argument components (e.g., premise) and then segmenting those components to their respective types \cite{stab2017parsing}. On the contrary, in this work, we identify tokens that represent claims as well as premises in one step from the original essays.
 
%we identify argument components such as \emph{claim} and \emph{premise} at the token level from essays written by school students.
 
% three claims (``Diet soda \dots the calories'', ``artificial sweeteners \dots effects, and ``these new \dots health'') appear to be present in the excerpt where only the third claim is supported by a premise (``there are \dots replaced'').

%explanation of the example 

We propose a detailed experimental setup to identify the argument components using both feature-based machine learning techniques and deep learning models. For the former, we used several structural, lexical, and syntactic features in a sequence classification framework using the Conditional Random Field (CRF) classifier \cite{lafferty2001conditional}. For the latter,  we employ a BiLSTM network and, finally, a transformer architecture - BERT \cite{devlin2019bert} with its pretrained and task-specific fine-tuned models. We achieve the best result from a particular BERT architecture (7.5\% accuracy improvement over the discrete features) that employs a joint multitask learning objective with an uncertainty-based weighting of two task-specific losses: (a) the \emph{main task} of token-level sequence classification, and (b) the \emph{auxiliary task} of sentence classification (i.e., whether a sentence contains argument or not). We make the dataset (student essays) from our research publicly available.\footnote{\href{https://github.com/EducationalTestingService/argument-component-essays}{https://github.com/EducationalTestingService/argument-component-essays}}

\section{Related Work}
\label{sec:rel_work}
The majority of the prior work on argument mining addressed the problems of argument segmentation, component, and relation identification modeled in a pipeline of subtasks \cite{peldszus-stede-2015-joint, stab2017parsing,potash2017here, niculae-etal-2017-argument} except a few research \cite{schulz2019challenges}. However, most of the research assumes the availability of segmented argumentative units and do the subsequent tasks such as the classification of argumentative component types \cite{biran2011identifying,stab2014identifying, park-cardie-2014-identifying}, argument relations \cite{ghosh-etal-2016-coarse,nguyen2016context}, and argument schemes \cite{hou-jochim-2017-argument, feng-hirst-2011-classifying}. 

Previous work on argument segmentation includes approaches that model the task as a sentence classification to argumentative or non-argumentative sentences \cite{moens2007automatic,Palau2009argumentation, mochales2011argumentation, rooney2012applying, lippi2015context, daxenberger2017essence,ajjour2017unit,petasis-2019-segmentation, chakrabarty-etal-2019-imho}, or by defining heuristics to identify argumentative segment boundaries \cite{madnani-etal-2012-identifying, persing2015modeling, persing-ng-2016-end, al-khatib-etal-2016-news}. Although we conduct segmentation, we focus on the token-level classification to directly identify the argument component's type. This setup is related to \newcite{schulz-etal-2018-multi} where authors analyzed students' diagnostic reasoning skills via token level identification. Our joint model using BERT is similar to \cite{eger2017neural}. However, we set the main task as the token-level classification where the auxiliary task of argumentative sentence identification assists the main task to attain a better performance.  
%sentence vs. paragraph in eger2017neural
%\newcite{ghosh-etal-2020-exploratory} mentioned that the recent surge in AI-informed systems has enabled assessing argumentation skills of students. 

Our research is based on essays written by middle school students. As stated earlier, most of the research on argumentative writing in an educational context focuses on identifying argument structures (i.e., argument components and their relations) \cite{persing2016end,nguyen2016context} as well as to predict essays scores from features derived from the essays (e.g.,   number  of claims and premises,   number  of  supported  claims, number of dangling claims) \cite{ghosh-etal-2016-coarse}. Related investigations have also examined the challenge of scoring a certain dimension of essay quality, such as relevance to the prompt \cite{persing2014modeling},  opinions and their targets \cite{farra-somasundaran-burstein:2015:bea}, argument strength \cite{persing2015modeling} among others. Majority of the above research are conducted in the context of college-level writing. For instance, \newcite{nguyen2018argument}  investigated argument structures in TOEFL11 corpus \cite{blanchard2013toefl11} which was also the main focus of \cite{ghosh-etal-2016-coarse}. \newcite{beigman_klebanov_detecting_2017} and \newcite{persing2015modeling} analyzed writing of university students and \newcite{stab2017parsing} used data from ``essayforum.com'', where college entrance examination  is the largest forum. Although, writing quality in essays by young writers has been addressed \cite{Attali_Burstein_2006,attali2008developmental,deane2014}, identification of arguments was not part of these studies. Computational analysis of arguments from school students is in infancy except for a few research \cite{lugini2018annotating,afrin-etal-2020-annotation,ghosh2020exploratory}. We believe our dataset (Section \ref{sec:data}) will be useful for researchers working at the intersection of argument mining and education.

\iffalse

\\
\\ \cite{nguyen-litman-2015-extracting}
\\ \cite{ajjour2017unit}
\\ \cite{eger2017neural}
\\ \cite{petasis-2019-segmentation}
\\
% does segmentation of argumentative tokens
% Argumentation and setup of the claim detection task (token, sentence classification)
% \\ nguyen - litman, (multiple work) 
% \\ stab - gurevych 
\\ recent work (Esin Durmus, etc.)
% Sequence Tagging
\\ deep learning, BERT 
\\ \cite{ghosh-etal-2020-exploratory}
% unit segmentation of argumentative texts [ajjour et al ; LSTM]
% segmentation of argumentative texts with contextualised word repr. [ petasis; biLSTM CRF]

\fi

\section{Data}
\label{sec:data}

% \debanjan{I will handle this section.}

\begin{table*}
    \centering
            \scalebox{0.8}{

    \begin{tabular}{l| l c|c c c c c|c}
        \hline\hline
        Corpora & Split & Essays & B-Claim & I-Claim &B-Premise &I-Premise &O-Arg &Total\\
        \hline
          & $training$ &100 &1,780 &21,966 &317 & 3,552 &51,478 &79,093\\
         $ARG2020$ & $dev$ &10 &171 &1,823 &32 &371 &4,008 &6,405\\
          & $test$ &35 &662 &8,207 &92 & 1,018 &14,987 &24,966\\
       %  \hline
        % $NARR2020$ & $test$ &20 &82 &991 &14 &118 &12,698 &13,903\\
        \hline\hline
    \end{tabular}
    }
    \caption{Token counts of each category in the $training$, $dev$, and $test$ sets of \emph{ARG2020}}
    \label{tab:data}
\end{table*}

We obtained a large number of English essays (over 10K) through the \emph{Writing Mentor} Educational App. This App is a Google Docs add-on designed to provide instructional writing support,  especially for academic writing. The add-on provides students to write argumentative or narrative essays and receive feedback on their writings. We selected a subset of 145 argumentative essays for the annotation purpose. Essays were either self-labeled as ``argumentative'' or annotators identified their argumentative nature from the titles (e.g., ``Should Artificial Sweeteners be Banned in America ?'').\footnote{Other metadata reveal that middle school students write these essays. However, we did not use any such information while annotating the essays.} Essays covered various social issues related to climate change, veteran care, effects of wars, whether sharks are dangerous or not, etc. We denote this corpus as $ARG2020$ in the remaining sections of the paper. We employed three expert annotators (with academic and professional background in Linguistics and Education) to identify the argument components. The annotators were instructed to read sentences from the essays and identify the \emph{claims} (defined as, ``a potentially arguable statement that indicates a person is arguing for or arguing against something. Claims are not clarification or elaboration statements.'') that the argument is in reference to. Next, once the claims are identified, the annotators annotated the \emph{premises} (defined as, ``reasons given by either for supporting or attacking the claims making those claims more than mere assertions'').\footnote {Definitions are from \cite{stab2017parsing} and Argument: Claims, Reasons, Evidence - Department of Communication, University of Pittsburgh (\href{https://bit.ly/396Ap3H}{https://bit.ly/396Ap3H})}  Earlier research has addressed college level writing, and even such resources are scarce except for a few corpora
\cite{stab2017parsing} 
(denoted as $SG2017$ in this paper). On the contrary, $ARG2020$ is based on middle school students writing, which differs from college level writing $SG2017$ in several aspects briefly discussed in the next paragraph.

First, we notice that essays in $SG2017$  maintain distinct paragraphs such as the introduction (initiates the major claim in the essay), the conclusion (summarizes the arguments), and a few paragraphs in between that express many claims and their premises. However, essays written by middle school students do not always comply with such writing conventions to keep a concrete introduction and conclusion paragraph, rather, they write many short paragraphs (7-8 paragraphs on average) per essay while each paragraph contains multiple claims. Second, in general, claims in college essays in $SG2017$ are justified by one or multiple premises, whereas $ARG2020$ has many unsupported claims. For instance, the excerpt from the annotated essay in Figure \ref{figure:tagged_intro} contains two unsupported claims (e.g., ``Diet soda, sugar \dots without the calories'' and ``artificial sweeteners \dots health effects''). Third, middle school students often put opinions (e.g., ``Sugar substitutes produce sweet food without feeling guilty consequences'') or matter-of-fact statements (e.g., ``Even canned food and dairy products can be artificially sweetened'') that are not argumentative claims but structurally they are identical to claims. Fourth, multiple claims frequently appear in a single sentence that are separated by discourse markers or commas. Fifth, many essays contain run-on sentences (e.g., ``this is hard on the family, they have a hard time adjusting'') that make the task of parsing even tricky. We argue these reasons make identifying argument claims and premises from $ARG2020$ more challenging. 

%fifth - many claims in one sentence. 
%Fourth, frequently, multiple claims appear in a single sentence that are separated by discourse markers or commas. 

The annotators were presented with specific guidelines and examples for annotation. We conducted a pilot task first where all the three annotators annotated ten essays and exchanged their notes for calibration. Following that, we continued pair-wise annotation tasks (30 essays for each pair of annotators), and finally, individual annotators annotated the remaining essays. Since the annotation task involves identifying each argumentative component's words, we have to account for fuzzy boundaries (e.g., in-claim vs. not-in-claim tokens) to measure the IAA. We considered the Krippendorff's $\alpha$ \cite{krippendorff2004measuring} metric to compute the IAA. We measure the $\alpha$ between each pair of annotators and report the average. For \emph{claim} we have a modest agreement of 0.71 that is comparable to \cite{stab2014identifying} and for \emph{premise}, we have a high agreement of 0.90.

%Here,   character overlap between any two annotations and the gap between them is utilized to measure the expected disagreement and the observed disagreement, respectively. 

%premmise -  and 0.66.    

%two IAA metrics usually used in literature for such cases: the information retrieval (IR) inspired precision-recall (P/R/F1) measure \cite{wiebe2005annotating} as well as Krippendorff' $\alpha$ \cite{krippendorff2004measuring}.  

%(Wiebe et al.,2005) and Krippendorff’s α (Krippendorff, 2004).We present here the main results; a detailed discussion of the IAA is left for a different paper. Following Wiebe et al. (2005), to calculate P/R/F1 for two annotators, one annotator’s ADUs are selected

Out of the 145 essays from $ARG2020$ we randomly assign 100 essays for  $training$, 10 essays for $dev$, and the remaining 35 essays for $test$. Table \ref{tab:data} represents the data statistics in the standard BIO format. 
%We also notice beside the ``argumentative'' essays middle-school students often submit ``narrative'' essays to the Educational App describing narrative stories. Although the majority of these narrative essays do not contain arguments, some essays do contain a few argumentative claims. Thus, we annotated a small set of 20 narrative essays as another test set (henceforth, denoted as $NARR_{test}$) for evaluation purpose. Two annotators (from the set of three annotators as described before) annotated the  $NARR_{test}$ essays. 
 %As expected, $ARG2020$ corpus contains much more arguments than $NARR_{test}$ corpus. 
We find the number of claims is almost six times the number of premises showing that the middle school students often fail to justify their proposed claims. We keep identifying opinions and argumentative relations (support/attack) as future work. 

\section{Experimental Setup}
\label{sec:exp_setup}
% \debanjan{create one section as Experimental Setup and then split that into (a) LR/CRF based discrete features (b) BERT, etc.}
Majority of the argumentation research  first segment the text in argumentative and non-argumentative segments and then identify the structures such as components and relations \cite{stab2017parsing}. \newcite{petasis-2019-segmentation} mentioned that the granularity of computational approaches addressing the second task of argument component identification is diverse because some approaches consider detecting components at the clause level (e.g., approaches focused on the $SG2017$ corpus \cite{stab2014identifying,stab2017parsing,ajjour2017unit,eger2017neural}) and others at the sentence levels \cite{chakrabarty-etal-2019-imho,daxenberger2017essence}. We avoided both approaches for the following two reasons. First, middle school student essays often contain run-on sentences, and it is unclear how to handle clause level annotations because parsing might be inaccurate. Second, around 62\% of the premises in the $training$ set appears to be in the same sentence as their claims. This makes sentence classification to either claim or premise impractical (Figure \ref{figure:tagged_intro} contains one such example). Thus, instead of relying on the pipeline approach, we tackle the problem by identifying argument components from the token-level classification akin to \newcite{schulz2019challenges}. Our unit of sequence tagging is a sentence, unlike a passage \cite{eger2017neural}. We apply a five-way token-classification (or sequence tagging) task while using the standard BIO notation for the claim and premise tokens (See Table \ref{tab:data}). Any token that is not ``B-Claim'', ``I-Claim'', ``B-Premise'', or ``I-Premise'' is denoted as ``O-Arg''. As expected, the number of ``O-Arg'' tokens is much larger than the other categories (see Table \ref{tab:data}). 
%example of claim/premise on same sentence. 

%In the first approach, we propose a three-way token-classification task while using the standard BIO notation (``B-Arg'', ``I-Arg'', ``O-Arg'') similar to many token segmentation research \cite{stab2017parsing}.

We explore three separate machine learning approaches well-established for studying token-based classification. First, we experiment with the sequence classifier Conditional Random Field (CRF)  that exploits  state-of-the-art discrete features. Second, we implement a BiLSTM network (with and without CRF) based on the BERT embeddings. Finally, we experiment with the fine-tuned BERT models with/without multitask learning setting. 
   
\subsection{Feature-based Models}
\label{ssec:crf-model}
Akin to \cite{stab2017parsing} we experiment with three groups of discrete features: \emph{structural}, \emph{syntactic} and \emph{lexico-syntactic} with some modifications. In addition, we experiment with embedding features extracted from the contextualized pre-trained language model of BERT.

%the sklearn-crfsuite \footnote{\href{https://sklearn-crfsuite.readthedocs.io}{https://sklearn-crfsuite.readthedocs.io}} 

% We test the aforementioned features in two classification approaches. First, token-based classification using the logistic regression classifier in the scikit-learn library \cite{scikit-learn} with grid search and feature group testing using the SKLL library.\footnote{\href{https://skll.readthedocs.org}{https://skll.readthedocs.org/}} Second, a sequence-based classification of tokens using a Conditional Random Field (CRF) classifier \cite{lafferty2001conditional} using the sklearn-crfsuite library.

\paragraph{Discrete Features}
For each token in a given essay, we extract structural features that include token position features and punctuation features. This set includes: whether the token is in the introduction (i.e., first-paragraph) or conclusion (i.e., last-paragraph) of the essay; whether the token is the first or the last in the sentence (excluding punctuations); the relative and absolute position of the token in document, paragraph and sentence; and the relative and absolute position of the sentence containing the token at the document and paragraph levels. The paragraph splits are determined by empty lines between sentences. The punctuation features include: whether the token is any punctuation or a period; and weather the preceding or succeeding token is any punctuation, a period, a comma or a semicolon. Such position features have shown to be useful in identifying claims and premises against sentences that do not contain any argument \cite{stab2017parsing}. We also extract syntactic features for each token that include part-of-speech tag of the token and normalized length to the lowest common ancestor (LCA) of the token and its preceding (and succeeding) token in the parse tree. In contrast with \cite{stab2017parsing}, we use dependency parsing as the base for the syntactic features rather than constituency parsing. Finally, we extract lexico-syntactic features (denoted as $lexSyn$ in Table \ref{tab:results_main}) that include the dependency relation governing the token in the dependency parse tree and the token itself, plus its governing dependency relation as another feature. This is also different than \cite{stab2017parsing} where the authors used lexicalized-parse tree \cite{collins2003head} to generate their lexico-syntactic features. These features are effective in identifying the argumentative discourse units \cite{stab2017parsing}. We also observed that using dependency parse trees as a basis for the lexico-syntactic features yields better results than constituency parse trees in our pilot experiments.
%TA: details below can be moved to an appendix
%The token position features include: whether the token is in the introduction (first-paragraph) or conclusion (last-paragraph) of the essay; whether the token is the first or the last in the sentence (excluding punctuations); the relative and absolute position of the token in document, paragraph and sentence; and the relative and absolute position of the sentence containing the token at the document and paragraph levels. The paragraph splits are determined by empty lines between sentences. The punctuation features include: whether the token is any punctuation or a period; and weather the preceding or succeeding token is any punctuation, a period, a comma or a semicolon. 
% In contrast to \cite{stab2017parsing}, we do not use consistency parse trees but rather dependency parse trees using their implementation in spaCy Python library.\footnote{\href{https://spacy.io/}{https://spacy.io/}} We also use the dependency relation between each token and its LCA token as another syntactic feature. Finally, we extract some lexico-syntactic features that include the dependency relation governing the token in the dependency parse tree and the token itself plus its governing dependency relation as another feature. This is also different than \cite{stab2017parsing} who use lexicalized-parse tree \cite{collins2003head} to generate their lexico-syntactic features. We found that using dependency parse trees as basis for the lexico-syntactic features to yield better results than constituency parse trees in our pilot experiments. 

\paragraph{Embedding Features from BERT}
BERT \cite{devlin2019bert}, a bidirectional transformer model, has achieved state-of-the-art performance in many NLP tasks. BERT is initially trained on the tasks of  masked language modeling (MLM) and next sentence prediction (NSP) over very large corpora of English Wikipedia and BooksCorpus. During its training, a special token ``[CLS]'' is added to the beginning of each training instance, and  the ``[SEP]'' tokens are  added to indicate the end of utterance(s) and separate, in case of two utterances. 

% In its standard form, 
Pretrained BERT (``bert-base-uncased'') can be used directly by extracting the token representations' embeddings. We use the average embeddings of the top four layers as suggested in \newcite{devlin2019bert}. For tokens with more than one word-piece when running BERT's tokenizer, their final embeddings feature is the average vector of all of their word-pieces. This feature yields a 768D-long vector that we use individually as well as in combination with the other discrete features in our experiments. We utilize the sklearn-crfsuite tool for our CRF experiments.\footnote{\href{https://sklearn-crfsuite.readthedocs.io}{https://sklearn-crfsuite.readthedocs.io}}

%for transfer learning by fine-tuning on a downstream task, i.e., argument relation identification where training instances are labeled as $Arg$ and $NoArg$. We denote this model as $BERT_{orig}$. BERT also allows \emph{adaptive pretraining} \cite{gururangan2020dont}, that is continued pretraining on an unlabeled corpus that can be task and domain relevant. We adaptively pretrained BERT over the entire $IAC$ corpus (excluding the $test$ partition), which consist of 11,800 unannotated discussion threads, consisting of 386,924 conversation turns. We denote this model as $BERT_{AP}$. In both the cases, BERT makes predictions via the ``[CLS]'' token.

\subsection{BiLSTM-CRF Models}
\label{ssec:bilstm}
To compare our models with standard sequence tagging models for argument segmentation \cite{petasis-2019-segmentation,ajjour-etal-2019-modeling, hua-etal-2019-argument}, we experiment with the BiLSTM-CRF sequence tagging model introduced by \citet{ma-hovy-2016-end} using the flair library \cite{akbik-etal-2019-flair}. We use the standard BERT (``bert-base-uncased'') embeddings (768D) in the embedding layer and projected to a single-layer BiLSTM of 256D. BiLSTMs provide the context to the token's left and right, which proved to be useful for sequence tagging tasks.  We train this model with and without a CRF decoder to see its effect on this task. The CRF layer considers both the output of the BiLSTM layer and the other neighboring tokens' labels, which improves the accuracy of the modeling desired transitions between labels \cite{ma-hovy-2016-end}.

\subsection{Transformers Fine-tuned Models}
\label{ssec:bert_finetunig}
Pre-trained BERT can also be used for transfer learning by fine-tuning on a downstream task, i.e., claim and premise token identification task where training instances are from the labeled dataset $ARG2020$. We denote this model as $BERT_{bl}$. Besides fine-tuning with the labeled data, we also experiment with a \emph{multitask learning} setting as well as conducted  
\emph{adaptive pretraining} \cite{gururangan2020dont}, that is continued pretraining on unlabeled corpora that can be task and domain relevant. We discuss the settings below.

\paragraph{Transformers Multitask Learning}

Multitask learning aims to leverage useful information in multiple related tasks to improve the performance of each task \cite{caruana1997multitask}. We treat the \emph{sequence labeling} task of five-way token-level
argument classification as the \emph{main} task while we adopt the binary task of \emph{sentence-level argument} identification (i.e., whether the candidate sentence contains an argument \cite{ghosh2020exploratory} as the \emph{auxiliary} task. Here, if any sentence in the candidate essay contains claim or premise token(s), the sentence is labeled as the positive category (i.e., argumentative), otherwise non-argumentative. We hypothesize that this auxiliary task of identifying argumentative sentences in a multitask setting could be useful for the main task of token-level classification.

We deploy two classification heads - one for each task (i.e., the main task of five-way token-level classification and the auxiliary task of identifying sentences containing arguments) and the relevant gold labels (size=5 and 2; \# of labels for the tasks, respectively) are passed to them. For the auxiliary task, the learned representation for the ``[CLS]'' token is passed to the classification head. The two losses from these individual heads are added and propagated back through the model. This allows BERT to model the nuances of both tasks and their interdependence simultaneously. However, instead of simply adding the losses from the two tasks, we employ \emph{dynamic weighting} of task-specific losses during the training process, based on the homoscedastic uncertainty of tasks, as proposed in \newcite{kendall2018multi}:
\begin{equation}
L =\sum_{t} \frac{1}{2\sigma^2_{t}}L_{t} + \log\sigma^2_{t}
\end{equation}
where $L_{t}$ and $\sigma_{t}$ depict the task-specific loss and its variance (updated through backpropagation), respectively, over the training instances. We denote this model as $BERT_{mt}$. 

\begin{figure}
    \hspace{0.3cm}
    \begin{framed}
  \includegraphics[width=0.9\linewidth]{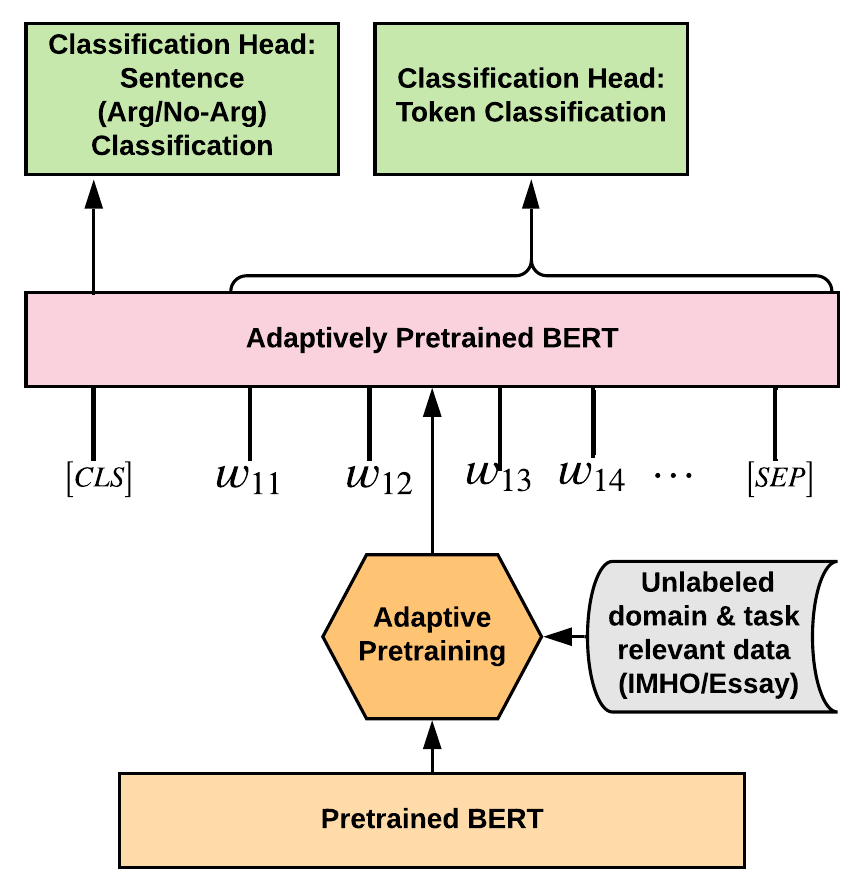}
    \end{framed}
\caption{BERT fine-tuning with adaptive pretraining on unlabeled data from a relevant domain followed by fine-tuning on the labeled dataset with the multitask variation.}
\label{figure:bertmodel}
\end{figure}

\paragraph{Adaptive Pre-training Learning}

We adaptively pretrained BERT over two unlabeled corpora. First, we train on a \emph{task relevant} Reddit corpus of 5.5 million opinionated claims that was released by \newcite{chakrabarty-etal-2019-imho}. These claims are self-labeled by the acronym: IMO/IMHO (in my (humble) opinion), which is commonly used in Reddit. We denote this model as $BERT_{IMHO}$. Next, we train on a \emph{task and domain relevant} corpus of around 10K essays that we obtained originally (See section \ref{sec:data}) from the \emph{Writing Mentor} App, excluding the annotated set of $ARG2020$ essays. We denote this model as $BERT_{essay}$. Figure \ref{figure:bertmodel} displays the use of the \emph{adaptive pretraining} step (in orange block) and the two classification heads (in green blocks) employed for the multitask variation.

For brevity, the parameter tuning description for all the models and experiments - discrete feature-based and deep-learning ones (e.g., CRF, BiLSTM, BERT) is in the supplemental material.

%feature extraction, we use it as a classier by fine-tuning on our labeled dataset at for the claim token detection task as shown in Figure \ref{fig:model1}, and by initially fine-tuning all layers on an unlabeled dataset from a relevant domain such as the IMHO corpus \cite{chakrabarty2019imho} and the GRE essays corpus \cite{ghosh-etal-2020-exploratory} as shown in Figure \ref{fig:model2}, which has been shown to improve the results on similar tasks in the cited previous work. Each of the unlabeled corpora (IMHO, GRE) have around 5 million sentences.

\begin{table*}[t]
    \centering
        \scalebox{0.8}{

    \begin{tabular}{l||c c c c c|c c}
        \hline \hline
          & \multicolumn{6}{c}{CRF} \\
         \hline
         Features &B-Claim &I-Claim &B-Premise &I-Premise &O-Arg &Acc. &F1\\
         \hline
         lexSyn                    &.395 &.530 &.114 &.176 &.768 &.673 &.397\\
         Discrete*                    &.269 &.504 &0 &.013 &.695 &.595 &.296\\
         Embeddings                      &.401 &.560 &.048 &.139 &.769 &.676 &.384\\
         Embeddings+lexSyn              &.482 &.610 &.134 &.180 &.764 &\underline{.682} &\underline{.434} \\
         Embeddings+Discrete*         &.434 &.593 &.055 &.152 &.762 &.676 &.399 \\

         \hline \hline
          & \multicolumn{6}{c}{BiLSTM} \\
         \hline
         Setup &B-Claim &I-Claim &B-Premise &I-Premise &O-Arg &Acc. &F1\\
         \hline
         BiLSTM             &.556 &.680 &.239 &.438 &.797 &\underline{.735} &\underline{.542} \\
         BiLSTM-CRF         &.558 &.676 &.199 &.378 &.789 &.727 &.520 \\
         
         \hline \hline
          & \multicolumn{6}{c}{BERT} \\
         \hline
         Setup &B-Claim &I-Claim &B-Premise &I-Premise &O-Arg &Acc. &F1\\
         \hline
         BERT$_{bl}$  &  .563 &  .674 & .274 & .425 & .795 & .728 & .546\\
        BERT$_{{bl}_{IMHO}}$   & .571  & .681 & .304 & .410 & .795 &  .730 & .540 \\
        BERT$_{{bl}_{essay}}$ & .564 &  .676 & .261 & .406 & .792 & \underline{\textbf{.747}} & \underline{{.561}} \\
        \hline
         BERT$_{mt}$  &  .567 &  .685 & .242 & .439 & .805 & .741 & .548\\
        BERT$_{{mt}_{IMHO}}$   & .562  & .684 & .221 & .413 & .794 &  .731 & .534\\
         BERT$_{{mt}_{essay}}$  &  .580 &  .702 & .254 & .427 & .810 & \underline{\textbf{.752}} & \underline{\textbf{.574}} \\%& \\
         \hline \hline
    \end{tabular}
    }
    \caption{F1 scores for Claim and Premise Token Detection on the test set. \underline{Underlined}: highest Accuracy/F1 in group. \textbf{Bold}: highest Accuracy/F1 overall. *Discrete: includes structural, syntactic, and lexSyn features.}
    \label{tab:results_main}
\end{table*}

\section{Results and Discussion}
\label{sec:results}
We present our experiments' results using the CRF, BiLSTM, and BERT models under different settings. We report the individual F1, Accuracy, and Macro-F1 (abbrev. as ``Acc.'' and ``F1'') scores for all the categories in Table \ref{tab:results_main} and Table \ref{tab:results_arg}.

We apply the discrete features (structural, syntactic, lexico-syntactic (``lexSyn''))  together and individually to the CRF model. We observe the structural and syntactic features do not perform well individually, especially in the case of premise tokens (See Table 5 %\ref{tab:results_discrete_appendix}
in Appendix A.3) and therefore, we only report the results of all discrete features (Discrete* in Table \ref{tab:results_main}) and individually only the performance of the lexSyn features. \newcite{stab2017parsing} noticed that structural features are effective to identify argument components, especially from the introduction and conclusion sections of the college level essays because they contain few argumentatively relevant content. On the contrary, as stated earlier, school student essays do not always comply with such writing conventions. Table \ref{tab:results_main} displays that the lexSyn feature independently performs better by almost 8\% accuracy than the combination of the other discourse features. This correlates to the findings from prior work on  $SG2017$ \cite{stab2017parsing} where the lexSyn features reached the highest F1 on a similar corpus. Next, we augment the embedding features from the BERT pre-trained model with the discrete features and notice a marginal improvement in the accuracy score (less than 1\%) over the performance of lexSyn features. This improvement is achieved from the higher accuracy in detecting the claim terms (e.g., Embedding+Discrete* achieves around 17\% and 10\%, an improvement over Discrete* features in the case of B-Claim and I-Claim, respectively). However, the accuracy of detecting the premise tokens (B-Premise and I-Premise) is still significantly low. We assume that this could be due to the low frequency of premises in the $training$ set, which seems to be more challenging for the CRF model to learn useful patterns from the pre-trained embeddings. On the contrary, the O-Arg token is the most frequent in the essays and that is reflected in the overall high accuracy scores for the O-Arg tokens (i.e., over 76\% on average).    

The overall performance(s) improve when we apply the BiLSTM networks on the $test$ data. Accuracy improves by 5.3\% in the case of BiLSTM against the Embeddings+lexSyn features (i.e., the best performance from the discrete feature sets). However, results do not improve when we augment the CRF classifier on top of the LSTM networks (BiLSTM-CRF). Instead, the performance drops by 0.8\% accuracy (See Table \ref{tab:results_main}). On related research, \newcite{petasis-2019-segmentation} have conducted extensive experiments with the BiLSTM-CRF architecture with various types of embeddings and demonstrated that only the specific combination of embeddings (e.g., GloVe+Flair+BERT) achieves higher performance than BiLSTM-only architecture. In the future, we are also interested in combining embeddings (e.g., GloVe with BERT). 
%for this line of experiments

%footnote -on petasis ls

In the case of BERT based experiments, we observe BERT$_{bl}$, i.e., the pre-trained BERT fine-tuned with the labeled $ARG2020$ corpus obtains an accuracy of 73\% that is comparable to the BiLSTM performance. In terms of the individual categories, we observe BERT$_{bl}$ achieves around 7.5\% improvement over the BiLSTM-CRF classifier for the B-Premise tokens. Moving on to the adaptive-pretrained models (e.g., BERT$_{IMHO}$ and BERT$_{essay}$), we observe both models perform better than the BERT$_{bl}$ where BERT$_{essay}$ achieves the best accuracy of 74.7\%, a 2\% improvement over BERT$_{bl}$. Although BERT$_{IMHO}$ was trained on a much larger corpus than BERT$_{essay}$ we assume since BERT$_{essay}$ was trained on a \emph{domain relevant} corpus (i.e., unlabeled student essays), it achieves the highest F1. Likewise, in the case of multitask models, we observe BERT$_{mt}$ performs better than BERT$_{bl}$ by 1.3\%. This shows that using argumentative sentence identification as an auxiliary task is beneficial for token-level classification. With regards to the adaptive-pretrained models, akin to the BERT$_{bl}$ based experiments, we observe BERT$_{mt_{essay}}$ perform best by achieving the highest accuracy over 75\%.

\paragraph{Argument Segmentation} We choose the five-way token-level classification of argument component over the standard pipeline approach because the standard level of granularity (i.e., sentence or clause-based) is not applicable to our $training$ data. Now, in order to test the benefit of the five-way token-level classification, we also compare it against the traditional approach of segmentation of argumentative units into argumentative and non-argumentative tokens. We again follow the standard BIO notation for a three-way token classification setup (B-Arg, I-Arg, and O-Arg) for argument segmentation. In this setup, the B-Claim and B-Premise classes are merged into B-Arg, and I-Claim and I-Premise are merged into I-Arg, while the O-Arg class remains unchanged. The results of all of our models on this task are shown in Table \ref{tab:results_arg}. We notice similar patterns (except for BERT$_{mt_{IMHO}}$ that performs better than BERT$_{mt}$ this time) in this three-way classification task as we saw in the five-way classification. The best model remains to be the BERT$_{mt_{essay}}$ with 77.3\% accuracy, which is an improvement of 2-3\% over the BiLSTM and other BERT-based architecture. 

In summary, we have two main observations from Table \ref{tab:results_main} and Table \ref{tab:results_arg}. First, the best model in Table \ref{tab:results_arg} reports only about 3\% improvement over the result from Table \ref{tab:results_main} which shows that the five-way token-level classification is comparable against the standard task of argument segmentation. Second, the accuracy of the argument segmentation task is much lower than the accuracy of college-level essay corpus $SG2017$ \cite{stab2017parsing} reported accuracy of 89.5\%). This supports the challenges of analyzing middle school student essays.

\begin{table}[t]
    \scalebox{0.8}{
    \begin{tabular}{l||c c c|c c}
        \hline \hline
          & \multicolumn{4}{c}{CRF} \\
         \hline
         Features               &B-Arg &I-Arg &O-Arg &Acc. &F1\\
         \hline
         lexSyn            &.385 &.518 &.768 &.683 &.557\\
         Discrete            &.288 &.493 &.710 &.625 &.497\\
         Embeddings              &.379 &.596 &.767 &.699 &.581\\
         Emb+lexSyn      &.468 &.622 &.768 &\underline{.708} &\underline{.619}\\
         Emb+Discrete     &.381 &.599 &.767 &.699 &.582\\
         
         \hline \hline
          & \multicolumn{4}{c}{BiLSTM} \\
         \hline
         Setup               &B-Arg &I-Arg &O-Arg &Acc. &F1\\
         \hline
         BiLSTM             &.546 &.730 &.792 & \underline{.759} &\underline{.689}\\
         BiLSTM-CRF         &.553 &.707 &.793 &.752 &.684\\
         
         \hline \hline
          & \multicolumn{4}{c}{BERT} \\
         \hline
         Setup                  &B-Arg &I-Arg &O-Arg &Acc. &F1\\
         \hline
         
         BERT$_{bl}$             & .567 & .698 & .795 &.750 & .687\\
          BERT$_{{bl}_{IMHO}}$   & .558 & .717 & .778 & .744 &.684\\
            BERT$_{{bl}_{essay}}$  & .567 & .707 & .795&  \underline{.754} & \underline{.690}\\
        \hline
                 BERT$_{mt}$    &   .555 & .702 & .803 & .758  & .688\\
         BERT$_{{mt}_{IMHO}}$    & .568 & .719 & .804 & .764 & .700\\
         BERT$_{{mt}_{essay}}$  & .563 & .735 & .811 & \textbf{\underline{.773}}  & \textbf{\underline{.710}}\\
         \hline \hline
    \end{tabular}}
    \caption{F1 scores for Argument Token Detection on the test set. \underline{Underlined}: highest Accuracy/F1 in group. \textbf{Bold}: highest Accuracy/F1 overall.}
    \label{tab:results_arg}
\end{table}

\subsection{Qualitative Analysis} \label{subsection:error}

Since we have explored three separate machine learning approaches with a variety of experiments, we analyze the results obtained from the BERT$_{{mt}_{essay}}$ model that has performed the best (Table \ref{tab:results_main}). According to the confusion matrix, there are three major sources of errors: (a) around 2500 ``O-Arg'' tokens are wrongly classified as ``I-Claim'' (b)  2162 ``I-Claim'' tokens are wrongly classified as  ``O-Arg'', and (c) 273 ``I-Premise'' tokens are erroneously classified as ``I-Claim''. Here, (a) and (b) are not surprising given these are the two categories with the largest number of tokens. For (c) we looked at a couple of examples, such as {``because of [Walmart 's goal of saving money]$_{premise}$, [customers see value in Walmart that is absent from other retailers]$_{claim}$''}. Here, the premise tokens are wrongly classified as O-Arg tokens. This is probably because the premise appears before the claim, which is uncommon in our $training$ set. We notice some of the other sources of errors, and we discuss them as follows:

\paragraph{non-arguments classified as arguments:} This error occurs often, but it is more challenging for \emph{opinions} or \emph{hypothetical examples} that resemble arguments but are not necessarily arguments. For instance, the opinion ``that actually makes me feel good afterward \dots'' and the hypothetical example ``Next , you will not be eating due to your lack of money'' are similar to an argument, and the classifier erroneously classifies them as claim.  In the future, we plan to include the labeled \emph{opinions} during training to investigate how the model(s) handle opinions vs. arguments during the classification.

\paragraph{missing multiple-claims from a sentence:} In many examples, we observe multiple claims appear in a single sentence, such as: {``[Some coral can recover from this]$_{claim}$ though [for most it is the final straw .]$_{claim}$''}. During prediction, the model predicts the first claim correctly but then starts the second claim with an ``I-Claim'' label, which is an impossible transition from ``Arg-O'' (i.e., does not enforce well-formed spans). Besides, the model starts the second claim wrongly at the word ``most'' rather than ``for''. This indicates the model's inability to distinguish discourse markers such as ``though'' as potential separators between argument components. This could be explained by the fact that markers such as ``though'' or ``because'' are frequently part of an argument claim. Such as, in {``[those games do not seem as violent even \underline{though} they are at the same level]$_{claim} $''}, ``though'' is labeled as ``I-Claim'. 

\paragraph{investigating run-on sentences:}
Some sentences contain multiple claims, which are written as one sentence via a comma-splice run-on such as {``[Humans in today 's world do not care about the consequences]$_{claim}$, [only the money they may gain .]$_{claim}$''} which has two claims in the gold annotations but it was predicted as one long claim by our best model. Another example is {``[The oceans are also another dire need in today's environment]$_{claim}$, each day becoming more filled with trash and plastics.''}, in which the claim is predicted correctly in addition to predicting an extra claim starting at the word {\it ``each''}. In the latter example, the model tends to over predicts claims when a comma comes in the middle of the sentence followed by a noun. However, in the former example, the adverb ``only'' that has a ``B-Claim'' label follows the comma rather than the more frequent nouns. Such instances add more complexity to understand and model argument structures in middle school student writing.

\paragraph{effect of the multitask learning:}
We examined the impact of multitask learning and notice two characteristics. First, as expected, the multitask  model can identify claims and premises that are missed by the single task model(s), i.e., the single task model missed the following claim: {``[many more negative effects that come with social media \dots'']$_{claim}$}'' that was correctly identified by the multitask model. Second, the clever handling of the back-propagation helps the multitask model to reduce false positives to be more precise. Many non-argumentative sentences, such as:  {``internet's social networks help teens find communities \dots''} and opinions, such as: {``take \$1.3 billion off \$11.3 billion the NCAA makes and give it to players''} are wrongly classified as claims by the single task models but are correctly classified as non-argumentative by the multitask model.

\section{Conclusion}
\label{sec:conc}

%Segmentation of argumentative units from text and subsequent identification of argument components are two major subtasks in argument mining research and they are usually applied as two separate tasks where one task (component identification) follows the first task (segmentation). On the contrary, instead of the pipeline approach 
We conduct a token-level classification task to identify the type of the argument component tokens (e.g., claims and premises) by combining the argument segmentation and component identification in one single task. We perused a new corpus collected from essays written by middle school students. We applied handcrafted discrete features, BiLSTM architecture, and transformers such as BERT under different conditions.  Our findings show that a multitask  BERT performs the best with an absolute gain of 7.5\% accuracy over the discrete features. We also conducted an in-depth comparison against the standard segmentation step (i.e., classifying the argumentative vs. non-argumentative units) and proposed a thorough qualitative analysis.

%fine-tuning with a language model tuned on five million unlabeled sentences from GRE essays has the best result on this task. This could lay the ground for applications such as automated writing assistance for students by highlighting all claims in an essay and showing claims that need more support if argument relations are modeled as well.
Middle school student essays often contain run-on sentences, multiple claims in a single sentence, or unsupported claims that make the task of identifying argument components much harder. We achieve the best performance using a multitask framework with an adaptive pretrained model, and we plan to continue  to augment other tasks (e.g., opinion and stance identification) under a similar multitask framework \cite{eger2017neural}. We can also generate personalized and relevant feedback for the students (e.g., which are the supported/unsupported claims in the essay?) that is useful in the paradigm of automated writing assistance.
%We also observe erroneous classification when the model(s) predicted I-claim tokens with no preceding B-claim token. A simple post-processing can fix such wrong transitions.  
Finally, we want to further experiment on another relevant task such as domain identification ({\sc SG2017} vs. {\sc ARG2020}) or genre prediction in essays (argumentative essays vs. narrative essays), which could be done on top of language model fine-tuning step on the relevant unlabeled corpus.

\section*{Acknowledgments}

The authors would like to thank Tuhin Chakrabarty, Elsbeth Turcan, Smaranda Muresan and Jill Burstein for their helpful comments and suggestions.  

\bibliography{anthology,eacl2021}

\begin{thebibliography}{60}
\expandafter\ifx\csname natexlab\endcsname\relax\def\natexlab#1{#1}\fi

\bibitem[{Afrin et~al.(2020)Afrin, Wang, Litman, Matsumura, and
  Correnti}]{afrin-etal-2020-annotation}
Tazin Afrin, Elaine~Lin Wang, Diane Litman, Lindsay~Clare Matsumura, and
  Richard Correnti. 2020.
\newblock \href {https://doi.org/10.18653/v1/2020.bea-1.7} {Annotation and
  classification of evidence and reasoning revisions in argumentative writing}.
\newblock In \emph{Proceedings of the Fifteenth Workshop on Innovative Use of
  NLP for Building Educational Applications}, pages 75--84, Seattle, WA, USA
  → Online. Association for Computational Linguistics.

\bibitem[{Ajjour et~al.(2019)Ajjour, Alshomary, Wachsmuth, and
  Stein}]{ajjour-etal-2019-modeling}
Yamen Ajjour, Milad Alshomary, Henning Wachsmuth, and Benno Stein. 2019.
\newblock \href {https://doi.org/10.18653/v1/D19-1290} {Modeling frames in
  argumentation}.
\newblock In \emph{Proceedings of the 2019 Conference on Empirical Methods in
  Natural Language Processing and the 9th International Joint Conference on
  Natural Language Processing (EMNLP-IJCNLP)}, pages 2922--2932, Hong Kong,
  China. Association for Computational Linguistics.

\bibitem[{Ajjour et~al.(2017)Ajjour, Chen, Kiesel, Wachsmuth, and
  Stein}]{ajjour2017unit}
Yamen Ajjour, Wei-Fan Chen, Johannes Kiesel, Henning Wachsmuth, and Benno
  Stein. 2017.
\newblock Unit segmentation of argumentative texts.
\newblock In \emph{Proceedings of the 4th Workshop on Argument Mining}, pages
  118--128.

\bibitem[{Akbik et~al.(2019)Akbik, Bergmann, Blythe, Rasul, Schweter, and
  Vollgraf}]{akbik-etal-2019-flair}
Alan Akbik, Tanja Bergmann, Duncan Blythe, Kashif Rasul, Stefan Schweter, and
  Roland Vollgraf. 2019.
\newblock \href {https://doi.org/10.18653/v1/N19-4010} {{FLAIR}: An easy-to-use
  framework for state-of-the-art {NLP}}.
\newblock In \emph{Proceedings of the 2019 Conference of the North {A}merican
  Chapter of the Association for Computational Linguistics (Demonstrations)},
  pages 54--59, Minneapolis, Minnesota. Association for Computational
  Linguistics.

\bibitem[{Al-Khatib et~al.(2016)Al-Khatib, Wachsmuth, Kiesel, Hagen, and
  Stein}]{al-khatib-etal-2016-news}
Khalid Al-Khatib, Henning Wachsmuth, Johannes Kiesel, Matthias Hagen, and Benno
  Stein. 2016.
\newblock \href {https://www.aclweb.org/anthology/C16-1324} {A news editorial
  corpus for mining argumentation strategies}.
\newblock In \emph{Proceedings of {COLING} 2016, the 26th International
  Conference on Computational Linguistics: Technical Papers}, pages 3433--3443,
  Osaka, Japan. The COLING 2016 Organizing Committee.

\bibitem[{Attali and Burstein(2006)}]{Attali_Burstein_2006}
Yigal Attali and Jill Burstein. 2006.
\newblock \href {https://ejournals.bc.edu/index.php/jtla/article/view/1650}
  {Automated essay scoring with e-rater v.2}.
\newblock \emph{The Journal of Technology, Learning and Assessment}, 4(3).

\bibitem[{Attali and Powers(2008)}]{attali2008developmental}
Yigal Attali and Don Powers. 2008.
\newblock A developmental writing scale.
\newblock \emph{ETS Research Report Series}, 2008(1):i--59.

\bibitem[{Beigman~Klebanov et~al.(2017)Beigman~Klebanov, Gyawali, and
  Song}]{beigman_klebanov_detecting_2017}
Beata Beigman~Klebanov, Binod Gyawali, and Yi~Song. 2017.
\newblock Detecting {Good} {Arguments} in a {Non-Topic-Specific} {Way}: {An}
  {Oxymoron}?
\newblock In \emph{Proceedings of the 55th {Annual} {Meeting} of the
  {Association} for {Computational} {Linguistics} ({Volume} 2: {Short}
  {Papers})}, pages 244--249, Vancouver, Canada. Association for Computational
  Linguistics.

\bibitem[{Biran and Rambow(2011)}]{biran2011identifying}
Or~Biran and Owen Rambow. 2011.
\newblock Identifying justifications in written dialogs.
\newblock In \emph{2011 IEEE Fifth International Conference on Semantic
  Computing}, pages 162--168. IEEE.

\bibitem[{Blanchard et~al.(2013)Blanchard, Tetreault, Higgins, Cahill, and
  Chodorow}]{blanchard2013toefl11}
Daniel Blanchard, Joel Tetreault, Derrick Higgins, Aoife Cahill, and Martin
  Chodorow. 2013.
\newblock Toefl11: A corpus of non-native english.
\newblock \emph{ETS Research Report Series}, 2013(2):i--15.

\bibitem[{Caruana(1997)}]{caruana1997multitask}
Rich Caruana. 1997.
\newblock Multitask learning.
\newblock \emph{Machine learning}, 28(1):41--75.

\bibitem[{Chakrabarty et~al.(2019)Chakrabarty, Hidey, and
  McKeown}]{chakrabarty-etal-2019-imho}
Tuhin Chakrabarty, Christopher Hidey, and Kathy McKeown. 2019.
\newblock \href {https://doi.org/10.18653/v1/N19-1054} {{IMHO} fine-tuning
  improves claim detection}.
\newblock In \emph{Proceedings of the 2019 Conference of the North {A}merican
  Chapter of the Association for Computational Linguistics: Human Language
  Technologies, Volume 1 (Long and Short Papers)}, pages 558--563, Minneapolis,
  Minnesota. Association for Computational Linguistics.

\bibitem[{Collins(2003)}]{collins2003head}
Michael Collins. 2003.
\newblock Head-driven statistical models for natural language parsing.
\newblock \emph{Computational linguistics}, 29(4):589--637.

\bibitem[{Correnti et~al.(2020)Correnti, Matsumura, Wang, Litman, Rahimi, and
  Kisa}]{correnti2020automated}
Richard Correnti, Lindsay~Clare Matsumura, Elaine Wang, Diane Litman, Zahra
  Rahimi, and Zahid Kisa. 2020.
\newblock Automated scoring of students’ use of text evidence in writing.
\newblock \emph{Reading Research Quarterly}, 55(3):493--520.

\bibitem[{Daxenberger et~al.(2017)Daxenberger, Eger, Habernal, Stab, and
  Gurevych}]{daxenberger2017essence}
Johannes Daxenberger, Steffen Eger, Ivan Habernal, Christian Stab, and Iryna
  Gurevych. 2017.
\newblock What is the essence of a claim? cross-domain claim identification.
\newblock In \emph{Proceedings of the 2017 Conference on Empirical Methods in
  Natural Language Processing}, pages 2055--2066.

\bibitem[{Deane(2014)}]{deane2014}
Paul Deane. 2014.
\newblock Using writing process and product features to assess writing quality
  and explore how those features relate to other literacy tasks.
\newblock \emph{ETS Research Report Series}, (1):1--23.

\bibitem[{Devlin et~al.(2019)Devlin, Chang, Lee, and
  Toutanova}]{devlin2019bert}
Jacob Devlin, Ming-Wei Chang, Kenton Lee, and Kristina Toutanova. 2019.
\newblock \href {https://doi.org/10.18653/v1/N19-1423} {{BERT}: Pre-training of
  deep bidirectional transformers for language understanding}.
\newblock In \emph{Proceedings of the 2019 Conference of the North {A}merican
  Chapter of the Association for Computational Linguistics: Human Language
  Technologies, Volume 1 (Long and Short Papers)}, pages 4171--4186,
  Minneapolis, Minnesota. Association for Computational Linguistics.

\bibitem[{Eger et~al.(2017)Eger, Daxenberger, and Gurevych}]{eger2017neural}
Steffen Eger, Johannes Daxenberger, and Iryna Gurevych. 2017.
\newblock Neural end-to-end learning for computational argumentation mining.
\newblock In \emph{Proceedings of the 55th Annual Meeting of the Association
  for Computational Linguistics (Volume 1: Long Papers)}, pages 11--22.

\bibitem[{Farra et~al.(2015)Farra, Somasundaran, and
  Burstein}]{farra-somasundaran-burstein:2015:bea}
Noura Farra, Swapna Somasundaran, and Jill Burstein. 2015.
\newblock Scoring persuasive essays using opinions and their targets.
\newblock In \emph{Proceedings of the Workshop on Innovative Use of NLP for
  Building Educational Applications}, pages 64--74.

\bibitem[{Feng and Hirst(2011)}]{feng-hirst-2011-classifying}
Vanessa~Wei Feng and Graeme Hirst. 2011.
\newblock \href {https://www.aclweb.org/anthology/P11-1099} {Classifying
  arguments by scheme}.
\newblock In \emph{Proceedings of the 49th Annual Meeting of the Association
  for Computational Linguistics: Human Language Technologies}, pages 987--996,
  Portland, Oregon, USA. Association for Computational Linguistics.

\bibitem[{Ghosh et~al.(2020)Ghosh, Beigman~Klebanov, and
  Song}]{ghosh2020exploratory}
Debanjan Ghosh, Beata Beigman~Klebanov, and Yi~Song. 2020.
\newblock An exploratory study of argumentative writing by young students: A
  transformer-based approach.
\newblock In \emph{Proceedings of the Fifteenth Workshop on Innovative Use of
  NLP for Building Educational Applications}, pages 145--150, Seattle, WA, USA.
  Association for Computational Linguistics.

\bibitem[{Ghosh et~al.(2016)Ghosh, Khanam, Han, and
  Muresan}]{ghosh-etal-2016-coarse}
Debanjan Ghosh, Aquila Khanam, Yubo Han, and Smaranda Muresan. 2016.
\newblock \href {https://doi.org/10.18653/v1/P16-2089} {Coarse-grained
  argumentation features for scoring persuasive essays}.
\newblock In \emph{Proceedings of the 54th Annual Meeting of the Association
  for Computational Linguistics (Volume 2: Short Papers)}, pages 549--554,
  Berlin, Germany. Association for Computational Linguistics.

\bibitem[{Ghosh et~al.(2014)Ghosh, Muresan, Wacholder, Aakhus, and
  Mitsui}]{ghosh2014analyzing}
Debanjan Ghosh, Smaranda Muresan, Nina Wacholder, Mark Aakhus, and Matthew
  Mitsui. 2014.
\newblock Analyzing argumentative discourse units in online interactions.
\newblock In \emph{Proceedings of the first workshop on argumentation mining},
  pages 39--48.

\bibitem[{Gururangan et~al.(2020)Gururangan, Marasovi{\'c}, Swayamdipta, Lo,
  Beltagy, Downey, and Smith}]{gururangan2020dont}
Suchin Gururangan, Ana Marasovi{\'c}, Swabha Swayamdipta, Kyle Lo, Iz~Beltagy,
  Doug Downey, and Noah~A. Smith. 2020.
\newblock Don{'}t stop pretraining: Adapt language models to domains and tasks.
\newblock In \emph{Proceedings of the 58th Annual Meeting of the Association
  for Computational Linguistics}, pages 8342--8360. Association for
  Computational Linguistics.

\bibitem[{Hou and Jochim(2017)}]{hou-jochim-2017-argument}
Yufang Hou and Charles Jochim. 2017.
\newblock \href {https://doi.org/10.18653/v1/W17-5107} {Argument relation
  classification using a joint inference model}.
\newblock In \emph{Proceedings of the 4th Workshop on Argument Mining}, pages
  60--66, Copenhagen, Denmark. Association for Computational Linguistics.

\bibitem[{Hua et~al.(2019)Hua, Hu, and Wang}]{hua-etal-2019-argument}
Xinyu Hua, Zhe Hu, and Lu~Wang. 2019.
\newblock \href {https://doi.org/10.18653/v1/P19-1255} {Argument generation
  with retrieval, planning, and realization}.
\newblock In \emph{Proceedings of the 57th Annual Meeting of the Association
  for Computational Linguistics}, pages 2661--2672, Florence, Italy.
  Association for Computational Linguistics.

\bibitem[{Kendall et~al.(2018)Kendall, Gal, and Cipolla}]{kendall2018multi}
Alex Kendall, Yarin Gal, and Roberto Cipolla. 2018.
\newblock Multi-task learning using uncertainty to weigh losses for scene
  geometry and semantics.
\newblock In \emph{Proceedings of the IEEE conference on computer vision and
  pattern recognition}, pages 7482--7491.

\bibitem[{Krippendorff(2004)}]{krippendorff2004measuring}
Klaus Krippendorff. 2004.
\newblock Measuring the reliability of qualitative text analysis data.
\newblock \emph{Quality and quantity}, 38:787--800.

\bibitem[{Lafferty et~al.(2001)Lafferty, McCallum, and
  Pereira}]{lafferty2001conditional}
John Lafferty, Andrew McCallum, and Fernando~CN Pereira. 2001.
\newblock Conditional random fields: Probabilistic models for segmenting and
  labeling sequence data.

\bibitem[{Lawrence and Reed(2020)}]{lawrence2020argument}
John Lawrence and Chris Reed. 2020.
\newblock Argument mining: A survey.
\newblock \emph{Computational Linguistics}, 45(4):765--818.

\bibitem[{Lippi and Torroni(2015)}]{lippi2015context}
Marco Lippi and Paolo Torroni. 2015.
\newblock Context-independent claim detection for argument mining.
\newblock In \emph{Twenty-Fourth International Joint Conference on Artificial
  Intelligence}.

\bibitem[{Lugini et~al.(2018)Lugini, Litman, Godley, and
  Olshefski}]{lugini2018annotating}
Luca Lugini, Diane Litman, Amanda Godley, and Christopher Olshefski. 2018.
\newblock Annotating student talk in text-based classroom discussions.
\newblock In \emph{Proceedings of the Thirteenth Workshop on Innovative Use of
  NLP for Building Educational Applications}, pages 110--116.

\bibitem[{Ma and Hovy(2016)}]{ma-hovy-2016-end}
Xuezhe Ma and Eduard Hovy. 2016.
\newblock \href {https://doi.org/10.18653/v1/P16-1101} {End-to-end sequence
  labeling via bi-directional {LSTM}-{CNN}s-{CRF}}.
\newblock In \emph{Proceedings of the 54th Annual Meeting of the Association
  for Computational Linguistics (Volume 1: Long Papers)}, pages 1064--1074,
  Berlin, Germany. Association for Computational Linguistics.

\bibitem[{Madnani et~al.(2012)Madnani, Heilman, Tetreault, and
  Chodorow}]{madnani-etal-2012-identifying}
Nitin Madnani, Michael Heilman, Joel Tetreault, and Martin Chodorow. 2012.
\newblock \href {https://www.aclweb.org/anthology/N12-1003} {Identifying
  high-level organizational elements in argumentative discourse}.
\newblock In \emph{Proceedings of the 2012 Conference of the North {A}merican
  Chapter of the Association for Computational Linguistics: Human Language
  Technologies}, pages 20--28, Montr{\'e}al, Canada. Association for
  Computational Linguistics.

\bibitem[{Mochales and Moens(2011)}]{mochales2011argumentation}
Raquel Mochales and Marie-Francine Moens. 2011.
\newblock Argumentation mining.
\newblock \emph{Artificial Intelligence and Law}, 19(1):1--22.

\bibitem[{Moens et~al.(2007)Moens, Boiy, Palau, and Reed}]{moens2007automatic}
Marie-Francine Moens, Erik Boiy, Raquel~Mochales Palau, and Chris Reed. 2007.
\newblock Automatic detection of arguments in legal texts.
\newblock In \emph{Proceedings of the 11th international conference on
  Artificial intelligence and law}, pages 225--230.

\bibitem[{Nguyen and Litman(2016)}]{nguyen2016context}
Huy Nguyen and Diane Litman. 2016.
\newblock Context-aware argumentative relation mining.
\newblock In \emph{Proceedings of the 54th Annual Meeting of the Association
  for Computational Linguistics (Volume 1: Long Papers)}, pages 1127--1137.

\bibitem[{Nguyen and Litman(2018)}]{nguyen2018argument}
Huy~V Nguyen and Diane~J Litman. 2018.
\newblock Argument mining for improving the automated scoring of persuasive
  essays.
\newblock In \emph{Thirty-Second AAAI Conference on Artificial Intelligence}.

\bibitem[{Niculae et~al.(2017)Niculae, Park, and
  Cardie}]{niculae-etal-2017-argument}
Vlad Niculae, Joonsuk Park, and Claire Cardie. 2017.
\newblock \href {https://doi.org/10.18653/v1/P17-1091} {Argument mining with
  structured {SVM}s and {RNN}s}.
\newblock In \emph{Proceedings of the 55th Annual Meeting of the Association
  for Computational Linguistics (Volume 1: Long Papers)}, pages 985--995,
  Vancouver, Canada. Association for Computational Linguistics.

\bibitem[{Palau and Moens(2009)}]{Palau2009argumentation}
Raquel~Mochales Palau and Marie-Francine Moens. 2009.
\newblock \href {https://doi.org/10.1145/1568234.1568246} {Argumentation
  mining: The detection, classification and structure of arguments in text}.
\newblock In \emph{Proceedings of the 12th International Conference on
  Artificial Intelligence and Law}, ICAIL '09, page 98–107, New York, NY,
  USA. Association for Computing Machinery.

\bibitem[{Park and Cardie(2014)}]{park-cardie-2014-identifying}
Joonsuk Park and Claire Cardie. 2014.
\newblock \href {https://doi.org/10.3115/v1/W14-2105} {Identifying appropriate
  support for propositions in online user comments}.
\newblock In \emph{Proceedings of the First Workshop on Argumentation Mining},
  pages 29--38, Baltimore, Maryland. Association for Computational Linguistics.

\bibitem[{Peldszus and Stede(2013)}]{peldszus2013argument}
Andreas Peldszus and Manfred Stede. 2013.
\newblock From argument diagrams to argumentation mining in texts: A survey.
\newblock \emph{International Journal of Cognitive Informatics and Natural
  Intelligence (IJCINI)}, 7(1):1--31.

\bibitem[{Peldszus and Stede(2015)}]{peldszus-stede-2015-joint}
Andreas Peldszus and Manfred Stede. 2015.
\newblock \href {https://doi.org/10.18653/v1/D15-1110} {Joint prediction in
  {MST}-style discourse parsing for argumentation mining}.
\newblock In \emph{Proceedings of the 2015 Conference on Empirical Methods in
  Natural Language Processing}, pages 938--948, Lisbon, Portugal. Association
  for Computational Linguistics.

\bibitem[{Persing and Ng(2014)}]{persing2014modeling}
Isaac Persing and Vincent Ng. 2014.
\newblock Modeling prompt adherence in student essays.
\newblock In \emph{Proceedings of the 52nd Annual Meeting of the Association
  for Computational Linguistics (Volume 1: Long Papers)}, pages 1534--1543.

\bibitem[{Persing and Ng(2015)}]{persing2015modeling}
Isaac Persing and Vincent Ng. 2015.
\newblock Modeling argument strength in student essays.
\newblock In \emph{Proceedings of the 53rd Annual Meeting of the Association
  for Computational Linguistics and the 7th International Joint Conference on
  Natural Language Processing (Volume 1: Long Papers)}, pages 543--552.

\bibitem[{Persing and Ng(2016{\natexlab{a}})}]{persing-ng-2016-end}
Isaac Persing and Vincent Ng. 2016{\natexlab{a}}.
\newblock \href {https://doi.org/10.18653/v1/N16-1164} {End-to-end
  argumentation mining in student essays}.
\newblock In \emph{Proceedings of the 2016 Conference of the North {A}merican
  Chapter of the Association for Computational Linguistics: Human Language
  Technologies}, pages 1384--1394, San Diego, California. Association for
  Computational Linguistics.

\bibitem[{Persing and Ng(2016{\natexlab{b}})}]{persing2016end}
Isaac Persing and Vincent Ng. 2016{\natexlab{b}}.
\newblock End-to-end argumentation mining in student essays.
\newblock In \emph{Proceedings of the 2016 Conference of the North American
  Chapter of the Association for Computational Linguistics: Human Language
  Technologies}, pages 1384--1394.

\bibitem[{Petasis(2019)}]{petasis-2019-segmentation}
Georgios Petasis. 2019.
\newblock \href {https://doi.org/10.18653/v1/W19-4501} {Segmentation of
  argumentative texts with contextualised word representations}.
\newblock In \emph{Proceedings of the 6th Workshop on Argument Mining}, pages
  1--10, Florence, Italy. Association for Computational Linguistics.

\bibitem[{Potash et~al.(2017)Potash, Romanov, and Rumshisky}]{potash2017here}
Peter Potash, Alexey Romanov, and Anna Rumshisky. 2017.
\newblock Here’s my point: Joint pointer architecture for argument mining.
\newblock In \emph{Proceedings of the 2017 Conference on Empirical Methods in
  Natural Language Processing}, pages 1364--1373.

\bibitem[{Rooney et~al.(2012)Rooney, Wang, and Browne}]{rooney2012applying}
Niall Rooney, Hui Wang, and Fiona Browne. 2012.
\newblock Applying kernel methods to argumentation mining.
\newblock In \emph{FLAIRS Conference}, volume 172.

\bibitem[{Schulz et~al.(2018)Schulz, Eger, Daxenberger, Kahse, and
  Gurevych}]{schulz-etal-2018-multi}
Claudia Schulz, Steffen Eger, Johannes Daxenberger, Tobias Kahse, and Iryna
  Gurevych. 2018.
\newblock \href {https://doi.org/10.18653/v1/N18-2006} {Multi-task learning for
  argumentation mining in low-resource settings}.
\newblock In \emph{Proceedings of the 2018 Conference of the North {A}merican
  Chapter of the Association for Computational Linguistics: Human Language
  Technologies, Volume 2 (Short Papers)}, pages 35--41, New Orleans, Louisiana.
  Association for Computational Linguistics.

\bibitem[{Schulz et~al.(2019)Schulz, Meyer, and
  Gurevych}]{schulz2019challenges}
Claudia Schulz, Christian~M Meyer, and Iryna Gurevych. 2019.
\newblock Challenges in the automatic analysis of students’ diagnostic
  reasoning.
\newblock In \emph{Proceedings of the AAAI Conference on Artificial
  Intelligence}, volume~33, pages 6974--6981.

\bibitem[{Somasundaran et~al.(2016)Somasundaran, Riordan, Gyawali, and
  Yoon}]{somasundaran2016evaluating}
Swapna Somasundaran, Brian Riordan, Binod Gyawali, and Su-Youn Yoon. 2016.
\newblock Evaluating argumentative and narrative essays using graphs.
\newblock In \emph{Proceedings of COLING 2016, the 26th International
  Conference on Computational Linguistics: Technical Papers}, pages 1568--1578.

\bibitem[{Song et~al.(2014)Song, Heilman, Klebanov, and
  Deane}]{song2014applying}
Yi~Song, Michael Heilman, Beata~Beigman Klebanov, and Paul Deane. 2014.
\newblock Applying argumentation schemes for essay scoring.
\newblock In \emph{Proceedings of the First Workshop on Argumentation Mining},
  pages 69--78.

\bibitem[{Stab and Gurevych(2014)}]{stab2014identifying}
Christian Stab and Iryna Gurevych. 2014.
\newblock Identifying argumentative discourse structures in persuasive essays.
\newblock In \emph{Proceedings of the 2014 Conference on Empirical Methods in
  Natural Language Processing (EMNLP)}, pages 46--56.

\bibitem[{Stab and Gurevych(2017)}]{stab2017parsing}
Christian Stab and Iryna Gurevych. 2017.
\newblock Parsing argumentation structures in persuasive essays.
\newblock \emph{Computational Linguistics}, 43(3):619--659.

\bibitem[{Stede and Schneider(2018)}]{stede2018argumentation}
Manfred Stede and Jodi Schneider. 2018.
\newblock Argumentation mining.
\newblock \emph{Synthesis Lectures on Human Language Technologies},
  11(2):1--191.

\bibitem[{Wacholder et~al.(2014)Wacholder, Muresan, Ghosh, and
  Aakhus}]{wacholder2014annotating}
Nina Wacholder, Smaranda Muresan, Debanjan Ghosh, and Mark Aakhus. 2014.
\newblock Annotating multiparty discourse: Challenges for agreement metrics.
\newblock In \emph{Proceedings of LAW VIII-The 8th Linguistic Annotation
  Workshop}, pages 120--128.

\bibitem[{Wachsmuth et~al.(2016)Wachsmuth, Al-Khatib, and
  Stein}]{wachsmuth-etal-2016-using}
Henning Wachsmuth, Khalid Al-Khatib, and Benno Stein. 2016.
\newblock \href {https://www.aclweb.org/anthology/C16-1158} {Using argument
  mining to assess the argumentation quality of essays}.
\newblock In \emph{Proceedings of {COLING} 2016, the 26th International
  Conference on Computational Linguistics: Technical Papers}, pages 1680--1691,
  Osaka, Japan. The COLING 2016 Organizing Committee.

\bibitem[{Zhang and Litman(2020)}]{zhang-litman-2020-automated}
Haoran Zhang and Diane Litman. 2020.
\newblock \href {https://doi.org/10.18653/v1/2020.acl-main.759} {Automated
  topical component extraction using neural network attention scores from
  source-based essay scoring}.
\newblock In \emph{Proceedings of the 58th Annual Meeting of the Association
  for Computational Linguistics}, pages 8569--8584, Online. Association for
  Computational Linguistics.

\end{thebibliography}
\bibliographystyle{acl_natbib}

% missing references
\begin{comment}
1. Schulz, C., Meyer, C. M., & Gurevych, I. (2019). Challenges in the Automatic Analysis of Students’ Diagnostic Reasoning. Proceedings of the AAAI Conference on Artificial Intelligence, 33(01), 6974-6981. https://doi.org/10.1609/aaai.v33i01.33016974
The paper discusses a different type of essays (very domain specific), but also frames the task of argument segmentation as a BIO sequence tagging task.

2. Luca Lugini, Diane Litman, Amanda Godley, Christopher Olshefski. Annotating Student Talk in Text-based Classroom Discussions. BEA 2018. https://www.aclweb.org/anthology/W18-0511/

3. Luca Lugini, Diane Litman. Contextual Argument Component Classification for Class Discussions. COLING 2020. https://www.aclweb.org/anthology/2020.coling-main.128/

5. Argumentation Mining in User-Generated Web Discourse. Ivan Habernal and Iryna Gurevych (Computational Linguistics, 2017)

6. Context Dependent Claim Detection" (Levy et al., COLING 2014)
\end{comment}

 \appendix
 % \newpage
\section{Appendix} \label{label:appendix}
\subsection{Parameter Tuning}
\label{tuning}

\paragraph{CRF experiment:}

For the { CRF model}, we search over the two regularization parameters c1 and c2 by sampling from exponential distributions with 0.5 scale for c1 and 0.05 scale for c2 using a 3 cross-validation over 50 iterations, which takes about 20 minutes of run-time. The final values are 0.8 for c1 and 0.05 for c2 for the best CRF model that uses LexSyn and BERT embeddings features. 

\paragraph{BiLSTM experiment:} 

For BiLSTM networks based experiments we searched the hyper parameters over the $dev$ set. Particularly we experimented with different mini-batch size (e.g., 16, 32), dropout value (e.g., 0.1, 0.3, 0.5, 0.7), number of epochs (e.g., 40, 50, 100 with early stopping), hidden state of sized-vectors (256). % and the Adam optimizer (learning rate of 0.01). 
Embeddings were generated using BERT (``bert-base-uncased'') (768 dimensions). After tuning we use the following hyper-parameters for the $test$ set: mini-batch size of 32, number of epochs = 100 (stop between 30-40 epochs), and dropout value of 0.1. The model has one BiLSTM layer with size 256 of the hidden layer.

\paragraph{BERT based models:}
We use the $dev$ partition for hyperparameter tuning (batch size of 8, 16, 32, 48), run for 3,5,6 epochs, learning rate of 3e-5) and optimized networks with the Adam optimizer. 
%The  BERT$_{TAP}$ was trained for 7 epochs and we saved the checkpoint(s) after each epoch. We use the final BERT$_{TAP}$ model (i.e., the model saved after 7 epochs).
The training partitions were  fine-tuned for 5 epochs with batch size = 16. Each training epoch took between 08:46 $\sim$ 9 minutes over a K-80 GPU with 48GB vRAM.

\subsection{Results of Discourse Feature Groups}
\label{ssec:discrete_res}
We show below the results of using each of the three feature groups individually: structural, syntactic and lexical-syntactic. As mentioned in the results section of the paper,  we can see below that the structural and syntactic features do not do well when used individually. Therefore, they were excluded from further experimentation with BERT embeddings.
\begin{table}
    \centering
        \scalebox{0.8}{
    \begin{tabular}{l||c c c c}
        \hline \hline
         Features &Structural &Syntactic &LexSyn &All Discrete\\
         \hline
            B-claim   &.294     &.218   &.395   &.269        \\
            I-claim   &.461   &.390    &.530    &.504         \\
            B-premise &0   &0   &.114  &0            \\
            I-premise &.018   &.009 &.176  &.013                \\
            O-Arg &.655 &.745 &.768 &.695 \\
            \hline
            Accuracy &.560 &.625 &.673 &.595 \\
            Macro F1 &.285   &.272     &.397      &.296 \\
         \hline \hline
    \end{tabular}
    }
    \caption{Accuracy and F1 scores for Claim and Premise Token Detection on the test set for each group of the discrete features in the CRF model.}
    \label{tab:results_discrete_appendix}
\end{table}

% \begin{table*}[t]
%     \centering
%         \scalebox{0.8}{

%     \begin{tabular}{l||c c c c c|c c}
%         \hline \hline
%           & \multicolumn{6}{c}{CRF} \\
%          \hline
%          Features &B-claim &I-claim &B-premise &I-premise &O-Arg &Acc. &F1\\
%          \hline
%          Structural                    &.294 &.461 &0 &.018 &.655 &.560 &.285\\
%          Syntactic                    &.218 &.390 &0 &.009 &.745 &.625 &.272\\
%          LexSyn                    &.395 &.530 &.114 &.176 &.768 &.673 &.397\\
%          All Discrete                    &.269 &.504 &0 &.013 &.695 &.595 &.296\\
%          \hline \hline
%     \end{tabular}
%     }
%     \caption{F1 scores for Claim and Premise Token Detection on the test set.}
%     \label{tab:results_discrete_appendix}
% \end{table*}

\end{document}